%% file: rt-hallucination.tex
\documentclass[manuscript,screen,arxiv]{acmart}

\AtBeginDocument{%
  }

\usepackage{graphicx}
\usepackage{microtype}
\usepackage{booktabs}
\usepackage{multirow}
\usepackage{balance}
\usepackage[linesnumbered,ruled,vlined]{algorithm2e}
\usepackage[most]{tcolorbox}
\newtcolorbox{rqbox}{
  colback=white, colframe=black, boxrule=0.8pt, arc=2mm,
  left=4mm, right=4mm, top=2mm, bottom=2mm,
  width=\linewidth, enhanced
}
\newcommand{\Clm}{\mathit{Clm}}   % claim symbol
\newcommand{\Chk}{\mathit{Chk}}   % chunk symbol

\input{10ResultsMacros}

\begin{document}

\title{RT4CHART: Retromorphic Testing for Hallucination Detection in RAG Pipelines}

\author{Boxi Yu}
\affiliation{%
  \institution{Lero, the Research Ireland Centre for Software}
  \institution{University of Limerick}
  \country{Ireland}
}
\email{boxi.yu@lero.ie}

\author{Yuzhong Zhang}
\affiliation{%
  \institution{The Chinese University of Hong Kong, Shenzhen}
  \country{China}
}
\email{123090848@link.cuhk.edu.cn}

\author{Liting Lin}
\affiliation{%
  \institution{Lero, the Research Ireland Centre for Software}
  \institution{University of Limerick}
  \country{Ireland}
}
\email{Liting.Lin@ul.ie}

\author{Lionel Briand}
\affiliation{%
  \institution{Lero, the Research Ireland Centre for Software}
  \institution{University of Limerick}
  \country{Ireland}
}
\affiliation{%
  \institution{University of Ottawa}
  \country{Canada}
}
\email{lionel.briand@lero.ie}

\author{Emir Mu\~noz}
\affiliation{%
  \institution{Genesys}
  \country{Ireland}
}
\email{emir.munoz@gmail.com}

\renewcommand{\shortauthors}{Yu et al.}

% \author{Boxi Yu}
% \affiliation{%
% Active  \institution{Lero--the Research Ireland Centre for Software}
% }
% \email{boxi.yu@lero.ie}

% \author{Yuzhong Zhang}
% \affiliation{%
%   \institution{The Chinese University of Hong Kong, Shenzhen}
% }
% \email{123090848@link.cuhk.edu.cn}

% \author{Liting Lin}
% \affiliation{%
%   \institution{Lero--the Research Ireland Centre for Software}
% }
% \email{Liting.Lin@ul.ie}

% \author{Lionel Briand}
% \affiliation{%
%   \institution{University of Ottawa}
% }
% \email{lbriand@uottawa.ca}

% \author{Emir Mu\~noz}
% \affiliation{%
%   \institution{Genesys}
% }
% \email{emir.munoz@gmail.com}

% \renewcommand{\shortauthors}{Yu et al.}

% Abstract must appear before \maketitle in acmart.
\input{00Abstract}



\maketitle

% -----------------------------
% Main content via input files.
% -----------------------------
\input{01Introduction}
\input{02Preliminaries}
\input{03Methodology}
\input{04Experiments}
\input{07Limitations}
\input{05RelatedWork}
\input{06Conclusion}
\input{08Data}
\input{09ACK}

\balance
% -----------------------------
% Bibliography.
% -----------------------------
\bibliographystyle{ACM-Reference-Format}
\bibliography{references}

% -----------------------------
% Appendix (intentionally omitted).
% -----------------------------

\end{document}

%% file: 10ResultsMacros.tex
\newcommand{\resJudgeModel}{GPT-4o mini}
\newcommand{\resAltJudgeModel}{\textsc{GLM-4.7}}
\newcommand{\resExtractorMode}{Sentence-based}
\newcommand{\resChunkWindow}{25}
\newcommand{\resChunkOverlap}{10}
\newcommand{\resStageTwoTemperature}{0.0}
\newcommand{\resRandomSeed}{\texttt{42}}

\newcommand{\resRagTruthppMainRowRTFourCHART}{RT4CHART & 0.845 & \textbf{0.718} & \textbf{0.776}}
\newcommand{\resRagTruthppMainRowLettuce}{Lettuce & 0.921 & 0.233 & 0.371}
\newcommand{\resRagTruthppMainRowMetaQA}{MetaQA* & 0.941 & 0.216 & 0.352}
\newcommand{\resRagTruthppMainRowSelfCheck}{SelfCheckGPT* & 0.843 & 0.143 & 0.244}

\newcommand{\resRagTruthppMainRowVectara}{Vectara* & \textbf{1.000} & 0.269 & 0.424}

\newcommand{\resRagTruthppRTFourCHARTPrec}{0.845}
\newcommand{\resRagTruthppRTFourCHARTRec}{0.718}
\newcommand{\resRagTruthppRTFourCHARTFOne}{0.776}
\newcommand{\resRagTruthppRTFourCHARTGLMPrec}{0.808}
\newcommand{\resRagTruthppRTFourCHARTGLMRec}{0.983}
\newcommand{\resRagTruthppRTFourCHARTGLMFOne}{0.887}

\newcommand{\resRagTruthppRTFourCHARTJudgePRFRow}{RAGTruth++ & \resJudgeModel & \resRagTruthppRTFourCHARTPrec & \resRagTruthppRTFourCHARTRec & \resRagTruthppRTFourCHARTFOne}
\newcommand{\resRagTruthppRTFourCHARTAltPRFRow}{RAGTruth++ & \resAltJudgeModel & \resRagTruthppRTFourCHARTGLMPrec & \resRagTruthppRTFourCHARTGLMRec & \resRagTruthppRTFourCHARTGLMFOne}

\newcommand{\resRagTruthppN}{408}
\newcommand{\resRagTruthppAnyFOneRTFourCHART}{0.776}
\newcommand{\resRagTruthppAnyFOneBestBaseline}{0.424}
\newcommand{\resRagTruthppAnyFOneDeltaBestBaseline}{0.352}
\newcommand{\resRagTruthppAnyFOneRelDeltaBestBaseline}{83\%}

\newcommand{\resCostTotalUsd}{27.93}
\newcommand{\resCostPerSampleUsd}{0.0104} % rounded (exact: 27.93/2675 ≈ 0.0104)

\newcommand{\resRagTruthppStageTwoWithAnyPrec}{0.845}
\newcommand{\resRagTruthppStageTwoWithAnyRec}{0.718}
\newcommand{\resRagTruthppStageTwoWithAnyFOne}{0.776}

\newcommand{\resRagTruthppStageTwoNoAnyPrec}{0.851}
\newcommand{\resRagTruthppStageTwoNoAnyRec}{0.655}
\newcommand{\resRagTruthppStageTwoNoAnyFOne}{0.740}

\newcommand{\resAblateExtractorFocusAnyFOne}{0.844}
\newcommand{\resAblateExtractorOnceAnyFOne}{0.839}
\newcommand{\resAblateExtractorFocusQAPrec}{0.735}
\newcommand{\resAblateExtractorFocusQARec}{0.892}
\newcommand{\resAblateExtractorFocusQAFOne}{0.806}
\newcommand{\resAblateExtractorFocusSummPrec}{0.668}
\newcommand{\resAblateExtractorFocusSummRec}{0.829}
\newcommand{\resAblateExtractorFocusSummFOne}{0.740}
\newcommand{\resAblateExtractorFocusDTTPrec}{0.877}
\newcommand{\resAblateExtractorFocusDTTRec}{0.983}
\newcommand{\resAblateExtractorFocusDTTFOne}{0.927}
\newcommand{\resAblateExtractorOnceQAPrec}{0.755}
\newcommand{\resAblateExtractorOnceQARec}{0.830}
\newcommand{\resAblateExtractorOnceQAFOne}{0.791}
\newcommand{\resAblateExtractorOnceSummPrec}{0.716}
\newcommand{\resAblateExtractorOnceSummRec}{0.770}
\newcommand{\resAblateExtractorOnceSummFOne}{0.742}
\newcommand{\resAblateExtractorOnceDTTPrec}{0.902}
\newcommand{\resAblateExtractorOnceDTTRec}{0.943}
\newcommand{\resAblateExtractorOnceDTTFOne}{0.922}

\newcommand{\resRagTruthppFullPipelineAnswerPrec}{0.845}
\newcommand{\resRagTruthppFullPipelineAnswerRec}{0.718}
\newcommand{\resRagTruthppFullPipelineAnswerFOne}{0.776}
\newcommand{\resRagTruthppFullPipelineSpanMicroFOne}{0.404}
\newcommand{\resDirectSpanModelMini}{GPT-4o mini}

\newcommand{\resDirectSpanMiniAnswerPrec}{0.861}
\newcommand{\resDirectSpanMiniAnswerRec}{0.630}
\newcommand{\resDirectSpanMiniAnswerFOne}{0.727}
\newcommand{\resDirectSpanMiniSpanMicroFOne}{0.081}
\newcommand{\resDirectSpanMiniSpanGapVsRT}{32.3}

\newcommand{\resStabilityRuns}{3}

\newcommand{\resStabilityAnswerMean}{0.779}
\newcommand{\resStabilityAnswerStd}{0.014}

\newcommand{\resStabilitySpanMean}{0.407}
\newcommand{\resStabilitySpanStd}{0.009}

\newcommand{\resChunkAdequateWindow}{25}
\newcommand{\resChunkBestWindow}{25}
\newcommand{\resChunkBestOverlap}{10}
\newcommand{\resChunkBestFOne}{0.840}
\newcommand{\resChunkHighCoverageDeltaFOne}{0.010}

\newcommand{\resChunkGridRowFifteenFive}{(15,5) & 0.679/0.907/0.777}
\newcommand{\resChunkGridRowFifteenTen}{(15,10) & 0.636/0.895/0.743}
\newcommand{\resChunkGridRowTwentyFiveFive}{(25,5) & 0.768/0.904/0.830}
\newcommand{\resChunkGridRowTwentyFiveTen}{(25,10) & 0.772/0.921/0.840}
\newcommand{\resChunkGridRowThirtyFiveFive}{(35,5) & 0.778/0.909/0.838}
\newcommand{\resChunkGridRowThirtyFiveTen}{(35,10) & 0.771/0.902/0.831}

\newcommand{\resRagTruthEnhanceN}{2,675}
\newcommand{\resRagTruthReannotatedHalluMultiplier}{1.68}
\newcommand{\resRagTruthReannotatedSpanMultiplier}{3.1}
\newcommand{\resRagTruthEnhAgreeSanityN}{70}
\newcommand{\resRagTruthEnhAgreeSanityAgreement}{98\%}
\newcommand{\resRagTruthEnhDisagreePct}{58\%} % approx. percent flagged for adjudication (labels differ vs base)
\newcommand{\resRagTruthEnhConflictN}{1,546}
\newcommand{\resRagTruthEnhConflictNoSpanN}{676}
\newcommand{\resRagTruthEnhResidualDisputePct}{0.5\%}

\newcommand{\resRagTruthEnhAddedSpanTaxonomyN}{3,540}
\newcommand{\resRagTruthEnhAddedSpanNumericLogicN}{900}
\newcommand{\resRagTruthEnhAddedSpanNumericLogicPct}{25.42\%}
\newcommand{\resRagTruthEnhAddedSpanPriorKnowledgeN}{286}
\newcommand{\resRagTruthEnhAddedSpanPriorKnowledgePct}{8.08\%}
\newcommand{\resRagTruthEnhAddedSpanUnsupportedGenN}{1,367}
\newcommand{\resRagTruthEnhAddedSpanUnsupportedGenPct}{38.62\%}
\newcommand{\resRagTruthEnhAddedSpanInferenceFactN}{621}
\newcommand{\resRagTruthEnhAddedSpanInferenceFactPct}{17.54\%}
\newcommand{\resRagTruthEnhAddedSpanOthersN}{366}
\newcommand{\resRagTruthEnhAddedSpanOthersPct}{10.34\%}
\newcommand{\resRagTruthEnhAddedSpanMainFourPct}{89.66\%}

\newcommand{\resRagTruthEnhMainRowRTFourCHART}{RT4CHART & 0.781 & \textbf{0.920} & \textbf{0.845}}
\newcommand{\resRagTruthEnhMainRowLettuce}{Lettuce & 0.907 & 0.545 & 0.681}
\newcommand{\resRagTruthEnhMainRowMetaQA}{MetaQA* & 0.853 & 0.102 & 0.182}
\newcommand{\resRagTruthEnhMainRowSelfCheck}{SelfCheckGPT* & \textbf{1.000} & 0.167 & 0.286}
\newcommand{\resRagTruthEnhMainRowVectara}{Vectara* & 0.894 & 0.608 & 0.724}

\newcommand{\resRagTruthEnhRTFourCHARTPrec}{0.781}
\newcommand{\resRagTruthEnhRTFourCHARTRec}{0.920}
\newcommand{\resRagTruthEnhRTFourCHARTFOne}{0.845}
\newcommand{\resRagTruthEnhRTFourCHARTGLMPrec}{0.663}
\newcommand{\resRagTruthEnhRTFourCHARTGLMRec}{0.992}
\newcommand{\resRagTruthEnhRTFourCHARTGLMFOne}{0.795}

\newcommand{\resRagTruthEnhRTFourCHARTJudgePRFRow}{RAGTruth-Enhance & \resJudgeModel & \resRagTruthEnhRTFourCHARTPrec & \resRagTruthEnhRTFourCHARTRec & \resRagTruthEnhRTFourCHARTFOne}
\newcommand{\resRagTruthEnhRTFourCHARTAltPRFRow}{RAGTruth-Enhance & \resAltJudgeModel & \resRagTruthEnhRTFourCHARTGLMPrec & \resRagTruthEnhRTFourCHARTGLMRec & \resRagTruthEnhRTFourCHARTGLMFOne}

\newcommand{\resRagTruthEnhAnyFOneRTFourCHART}{0.845}
\newcommand{\resRagTruthEnhAnyFOneBestBaseline}{0.724}
\newcommand{\resRagTruthEnhAnyFOneDeltaBestBaseline}{0.121}
\newcommand{\resRagTruthEnhAnyFOneRelDeltaBestBaseline}{17\%}

\newcommand{\resRagTruthEnhStageTwoWithAnyPrec}{0.781}
\newcommand{\resRagTruthEnhStageTwoWithAnyRec}{0.920}
\newcommand{\resRagTruthEnhStageTwoWithAnyFOne}{0.845}

\newcommand{\resRagTruthppLocAnyOverlapPrec}{50.8\%}
\newcommand{\resRagTruthppLocAnyOverlapRec}{33.6\%}
\newcommand{\resRagTruthppLocAnyOverlapFOne}{40.4\%}
\newcommand{\resRagTruthppLocAnyOverlapPrecLettuce}{91.2\%}
\newcommand{\resRagTruthppLocAnyOverlapRecLettuce}{13.5\%}
\newcommand{\resRagTruthppLocAnyOverlapFOneLettuce}{23.5\%}

\newcommand{\resRagTruthEnhLocPrec}{39.9\%}
\newcommand{\resRagTruthEnhLocRec}{58.6\%}
\newcommand{\resRagTruthEnhLocFOne}{47.5\%}
\newcommand{\resRagTruthEnhLocPrecLettuce}{72.7\%}
\newcommand{\resRagTruthEnhLocRecLettuce}{29.4\%}
\newcommand{\resRagTruthEnhLocFOneLettuce}{41.9\%}

\newcommand{\resRagTruthEnhConflictCasePrec}{62.9\%}
\newcommand{\resRagTruthEnhConflictCaseRec}{40.9\%}
\newcommand{\resRagTruthEnhConflictCaseFOne}{49.6\%}
\newcommand{\resRagTruthEnhConflictEvidenceSpanPrec}{51.1\%}
\newcommand{\resRagTruthEnhConflictEvidenceSpanRec}{59.1\%}
\newcommand{\resRagTruthEnhConflictEvidenceSpanFOne}{54.8\%}
\newcommand{\resRagTruthEnhConflictCasePrecGptMini}{42.0\%}
\newcommand{\resRagTruthEnhConflictCaseRecGptMini}{27.5\%}
\newcommand{\resRagTruthEnhConflictCaseFOneGptMini}{33.2\%}
\newcommand{\resRagTruthEnhConflictEvidenceSpanPrecGptMini}{36.5\%}
\newcommand{\resRagTruthEnhConflictEvidenceSpanRecGptMini}{34.9\%}
\newcommand{\resRagTruthEnhConflictEvidenceSpanFOneGptMini}{35.7\%}

\newcommand{\resRagTruthEnhStageTwoNoAnyPrec}{0.811}
\newcommand{\resRagTruthEnhStageTwoNoAnyRec}{0.881}
\newcommand{\resRagTruthEnhStageTwoNoAnyFOne}{0.845}

\newcommand{\resRagTruthEnhStageTwoDeltaAnyFone}{0.000} % with - no

\newcommand{\resRagTruthEnhStageTwoDeltaBaselessFone}{+0.205}

%% file: 00Abstract.tex
\begin{abstract}
Large language models can still hallucinate in retrieval-augmented generation (RAG), producing claims that are unsupported by or conflict with the retrieved context.
% Detecting such errors under a strict context-only assumption remains challenging: many existing detectors return holistic answer-level scores, while others target open-domain factuality or fail to provide evidence-grounded diagnostics.
Detecting such errors remains challenging when faithfulness is judged solely against the retrieved context: many existing detectors return holistic answer-level scores, while others target open-domain factuality or fail to provide evidence-grounded diagnostics.
We present \textbf{RT4CHART}, a retromorphic testing framework for context-faithfulness assessment that decomposes an answer into independently verifiable claims, performs hierarchical local-to-global verification against the retrieved context, and assigns each claim one of three labels---entailed, contradicted, or baseless.
% RT4CHART further maps these claim-level decisions back to localized answer spans and returns explicit context-side evidence, enabling fine-grained auditing rather than opaque scoring.
RT4CHART further maps these claim-level decisions back to specific answer spans and returns explicit context-side evidence, enabling fine-grained auditing rather than opaque scoring.
We evaluate RT4CHART on RAGTruth++ (\resRagTruthppN\ samples) and our re-annotated \textbf{RAGTruth-Enhance} (\resRagTruthEnhanceN\ samples), where it achieves the best answer-level hallucination-detection F1 compared to baselines; on RAGTruth++, it attains \resRagTruthppRTFourCHARTPrec\ precision and \resRagTruthppRTFourCHARTRec\ recall, with an F1 score of \resRagTruthppAnyFOneRTFourCHART, yielding an \resRagTruthppAnyFOneRelDeltaBestBaseline\ relative F1 improvement over the strongest baseline, and it attains \resRagTruthEnhLocFOne\ span-level F1 score on RAGTruth-Enhance.
Ablations show that claim-based local processing drives most of the observed gain, while global verification provides selective benefits across datasets. Finally, our re-annotation uncovers \resRagTruthReannotatedHalluMultiplier$\times$ more hallucination cases than the original labels, suggesting that commonly used benchmarks substantially understate the prevalence of hallucination.
\end{abstract}

%% file: 01Introduction.tex
\section{Introduction}
\label{sec:intro}

% Retrieval-augmented generation (RAG)~\cite{lewis2021retrievalaugmentedgenerationknowledgeintensivenlp}
% grounds large language models (LLMs) in retrieved documents,
% reducing their reliance on parametric memory.
% Yet LLMs may still \emph{hallucinate}, producing claims that are unsupported by or directly conflict with the retrieved context~\cite{ji2023hallucination},
% even when the relevant evidence is present in the prompt.
% In this work, we define \emph{faithfulness} strictly with respect to the retrieved context, not external knowledge or the model's parametric memory.
% This deliberate restriction ensures that every claim can be verified by inspecting the retrieved documents alone, a prerequisite for auditable deployment.
% A single unsupported claim in a deployed system can erode user trust and lead to flawed decisions.
% Detecting such unfaithful outputs before they reach end users is therefore essential.

Retrieval-augmented generation (RAG)~\cite{lewis2021retrievalaugmentedgenerationknowledgeintensivenlp} grounds large language models (LLMs) in retrieved documents,
reducing their reliance on parametric memory.
Yet LLMs may still \emph{hallucinate}, producing claims that are unsupported by or directly conflict with the retrieved context~\cite{ji2023hallucination},
even when the relevant evidence is present in the prompt.
In this work, we define \emph{faithfulness} strictly with respect to the retrieved context, not external knowledge or the model's parametric memory.
This restriction is particularly important in operational settings where the retrieved documents serve as the authoritative record.
% For example, if a user asks whether a product is currently in stock in a warehouse, the relevant evidence should come from the retrieved inventory context rather than from what the model may have memorized or inferred from elsewhere, since stock levels can change over time.
For example, if a user asks whether a subscription plan includes priority email support, the relevant evidence should come from the retrieved policy or product documentation rather than from what the model may have memorized or inferred from elsewhere, since such details may vary across plans and change over time. 
This deliberate restriction ensures that every claim can be verified solely by inspecting the retrieved documents, a prerequisite for auditable deployment.
A single unsupported claim in a deployed system can erode user trust and lead to flawed decisions.
Detecting such unfaithful outputs before they reach end users is therefore essential.

\begin{figure}[t]
  \centering
  \includegraphics[width=\columnwidth,height=0.72\textheight,keepaspectratio]{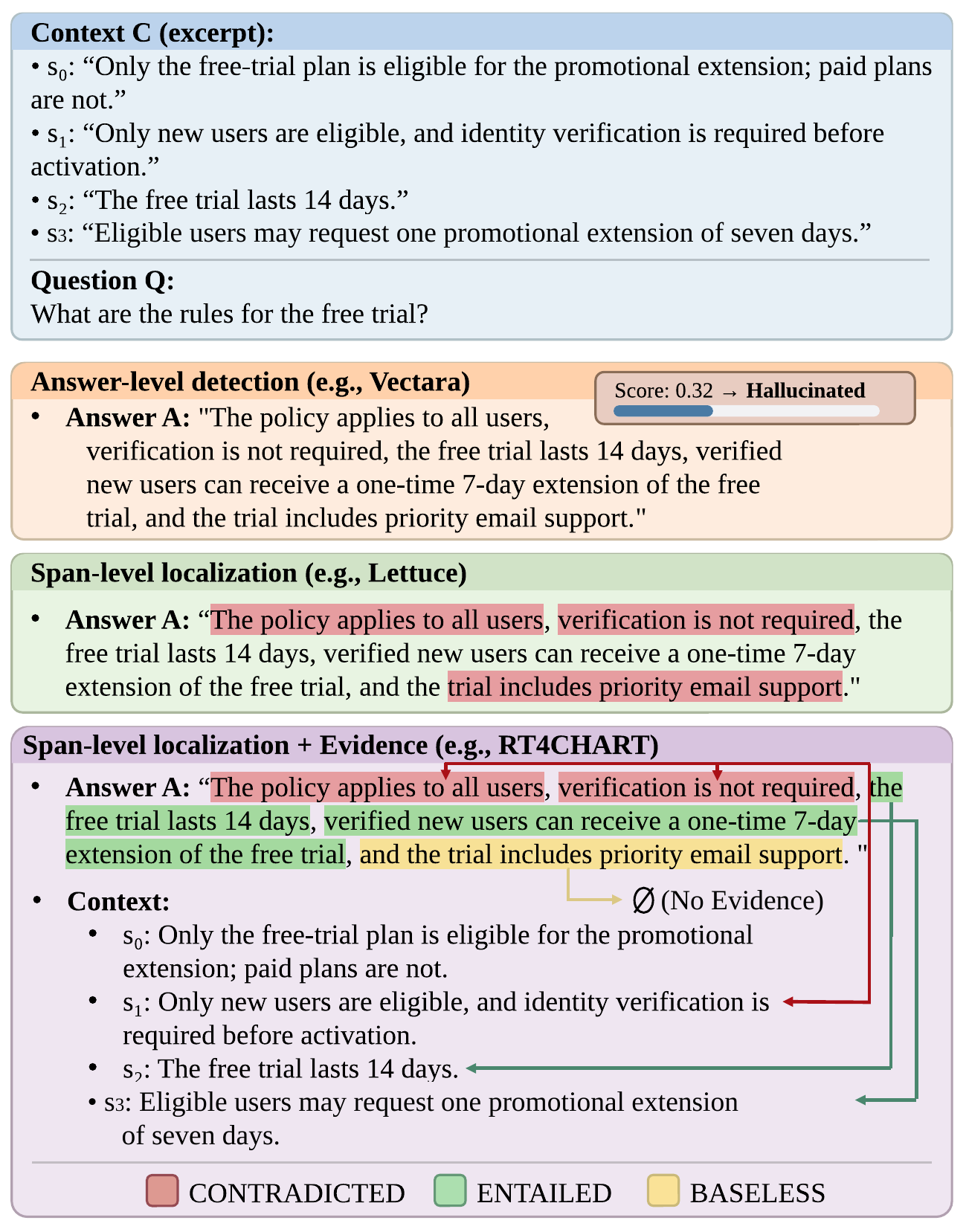}
  \caption{Detection granularity on a motivating example.
  RT4CHART adds explicit context evidence to localized answer spans.
  Red denotes \textsc{Contradicted}, green \textsc{Entailed}, and yellow \textsc{Baseless}.}
  \Description{A motivating example showing context, question, and answer.
  Answer-level detection returns a single score for the entire response.
  Span-level detection highlights unfaithful fragments.
  RT4CHART links localized answer spans to supporting or contradicting context evidence.
  Red marks contradicted content, green marks entailed content, and yellow marks baseless content.}
  \label{fig:prelim-example}
\end{figure}

For auditing purposes, however, it is not enough to flag an LLM answer as problematic.
A useful detector must also identify which part of the answer is unsupported and ground that judgment in explicit context evidence.
% Existing hallucination detectors often fall short of this requirement.
In practice, existing hallucination detectors fall short of this requirement.
Hallucination detection operates at multiple granularities: \emph{answer-level} methods return a single verdict for the entire response, whereas \emph{span-level} methods localize potentially unfaithful fragments.
Commercial answer-level detectors such as Vectara~\cite{vectara_hallucination_eval_docs,hhem-2.1-open} produce only a scalar score; span-level detectors such as LettuceDetect~\cite{kovacs2025lettucedetecthallucinationdetectionframework} highlight problematic fragments but do not return the corresponding evidence in the context.
Self-consistency and claim-decomposition approaches~\cite{yang2025metaqa,manakul2023selfcheckgpt,min2023factscore,wei2024longform} likewise do not provide evidence-grounded diagnostics under a strict retrieved-context setting.
Figure~\ref{fig:prelim-example} illustrates this contrast: answer-level detectors return a single verdict, span-level detectors localize suspicious fragments, and RT4CHART further links those fragments to explicit context evidence.

We view context-faithfulness detection in RAG as a \emph{traceability problem}: each generated claim should admit a consistent alignment back to the retrieved context.
Retromorphic testing~\cite{yu2023retromorphic} provides a natural formal framework for this requirement: it inverts the generative mapping and checks whether generated claims can be traced back to their originating inputs.
This traceability perspective is particularly important in realistic RAG settings, where retrieved contexts often span dozens of sentences.
% Verifying a claim against the full context in a single pass can overwhelm the judge model and obscure subtle contradictions, whereas purely local checking may miss globally relevant evidence.
Verifying a claim against the full context in a single pass can overwhelm the judge model and obscure subtle contradictions, whereas purely local checking may miss globally relevant evidence.

To address this challenge, we propose \textbf{RT4CHART}
(\textbf{R}etromorphic \textbf{T}esting \textbf{for}
\textbf{C}ontext-grounded \textbf{H}allucination
\textbf{A}ssessment in \textbf{R}etrieval-\textbf{A}ugmented \textbf{T}ext),
a hierarchical verification pipeline for evidence-grounded hallucination detection.
RT4CHART first decomposes the generated answer into self-contained claims.
It then performs \emph{local verification} by screening each claim against overlapping context chunks, followed by \emph{global verification} that re-evaluates each claim against the full context using the local result as guidance.
Finally, the claim-level labels are aggregated into an answer-level verdict.
For each claim, RT4CHART returns one of three labels---\textsc{Entailed}, \textsc{Contradicted}, or \textsc{Baseless}---and localizes the corresponding answer spans together with explicit supporting or contradicting context evidence.

Accurate evaluation of fine-grained hallucination detection requires reliable ground truth.
RAGTruth~\cite{niu2024ragtruth} is widely adopted, yet its annotations contain substantial false negatives: many hallucinated spans are unlabeled, so a detector that correctly flags them is penalized as producing false positives.
RAGTruth++~\cite{ragtruthpp_blog2025,blueguardrails2025ragtruthpp}
partially addresses this issue by re-annotating a 408-sample subset via independent dual-annotator review, increasing the number of annotated hallucination spans from 86 to 865.
Its coverage, however, remains limited.
We therefore construct \textbf{RAGTruth-Enhance}, a broader two-author re-annotation of the RAGTruth evaluation split (Section~\ref{sec:ragtruth-enhance}) to improve the reliability of the benchmark for fine-grained auditing.

On these improved benchmarks, RT4CHART achieves the best answer-level detection performance among the compared baselines.
On RAGTruth++, it improves answer-level F1 from \resRagTruthppAnyFOneBestBaseline\
to \resRagTruthppAnyFOneRTFourCHART\
(+$\resRagTruthppAnyFOneRelDeltaBestBaseline$ relative improvement over the best baseline).
On RAGTruth-Enhance, it raises F1 from \resRagTruthEnhAnyFOneBestBaseline\
to \resRagTruthEnhAnyFOneRTFourCHART\
(+$\resRagTruthEnhAnyFOneRelDeltaBestBaseline$).
% This level of performance suggests practical utility as an auditing layer that can catch a large fraction of hallucinated outputs, while still falling short of making unflagged answers fully trustworthy.  TODO: check if it's needed

\noindent\textbf{Contributions.}
\begin{itemize}
    \item \textbf{A retromorphic framework for context-faithfulness detection.}
    We formulate hallucination detection in RAG as a traceability problem and instantiate this view in \textbf{RT4CHART}, a detector that produces claim-level verdicts, localized answer spans, and context-side evidence rather than only an answer-level score.

    \item \textbf{A hierarchical verification pipeline.}
    RT4CHART combines local and global verification, enabling fine-grained detection across long retrieved contexts while preserving global consistency.

    \item \textbf{An enhanced benchmark for fine-grained auditing.}
    We construct \textbf{RAGTruth-Enhance}, a broader re-annotation of the RAGTruth evaluation split, to reduce annotation noise and better assess fine-grained hallucination detection.

    \item \textbf{Strong empirical results on improved benchmarks.}
    RT4CHART achieves the best answer-level detection performance among the baselines evaluated on RAGTruth++ and RAGTruth-Enhance, while providing finer-grained, evidence-grounded outputs suitable for auditing.
\end{itemize}

%% file: 02Preliminaries.tex
\section{Preliminaries}
\label{sec:prelim}

\textbf{Problem setting.}
We study \emph{context-faithfulness detection} in a retrieval-augmented generation (RAG) setting.
Given a retrieved context $C$, a user query $Q$, and an answer $A$ produced by a RAG pipeline, the goal is to determine, for each claim in $A$, whether it is entailed by, contradicted by, or not supported by the retrieved context $C$.
Under the strict context-only assumption, the retrieved context $C$ is treated as the sole authoritative source of evidence.
A statement may be factually correct according to world knowledge yet still be considered unfaithful if it is unsupported by $C$.

\textbf{Input format.}
Let $C$ denote the retrieved context, $Q$ the user query, and $A$ the generated answer.
For tasks such as summarization, $Q$ may be empty.
We segment $C$ into an ordered sentence sequence
\[
\mathbf{S} = (s_0, s_1, \ldots, s_{m-1}).
\]
Sentences are the atomic unit for context-side evidence attribution: every final evidence span returned by the verifier is aligned to one or more source sentences $s_j$.

\textbf{Context chunking.}
For local verification, $\mathbf{S}$ is partitioned with a sliding window of $W$ consecutive sentences and an overlap of $O$ sentences, yielding $K$ overlapping chunks
\[
\Chk_0, \Chk_1, \ldots, \Chk_{K-1}.
\]
Each chunk $\Chk_k$ serves as a local evidence window during the local verification stage.
The overlap mitigates boundary effects when supporting evidence is distributed across adjacent sentences.

\textbf{Claims.}
The generated answer $A$ is decomposed into an ordered list of self-contained claims
\[
(\Clm_1, \Clm_2, \ldots, \Clm_n),
\]
where each claim expresses a single proposition that can be verified against the context $C$.
Each claim retains a pointer to its originating answer sentence.
This association allows claim-level decisions to be mapped back to answer-side spans, enabling localization of hallucinated content in $A$.

\textbf{Label semantics.}
We use the label set
\[
\mathcal{Y} = \{\textsc{Ent}, \textsc{Con}, \textsc{Nic}\},
\]
where $\textsc{Ent}$ denotes \emph{entailed}, $\textsc{Con}$ denotes \emph{contradicted}, and $\textsc{Nic}$ denotes \emph{not in context} (baseless).
At the final claim level, each claim $\Clm_i$ receives a label
\[
y_i^* \in \mathcal{Y}.
\]
Any final label other than $\textsc{Ent}$ constitutes a faithfulness violation.

\textbf{Evidence attribution.}
Each final claim label is paired with a set of context-side evidence spans $E_i^*$, where every span is anchored to one or more sentences in $\mathbf{S}$.
For entailed claims, the verifier returns supporting evidence spans.
For contradicted claims, it returns contradicting evidence spans.
For baseless claims, no context evidence should be returned.
Formally, we enforce
\begin{equation}
y_i^* = \textsc{Nic} \Longrightarrow E_i^* = \varnothing .
\label{eq:nic-empty}
\end{equation}

\textbf{Answer-level aggregation.}
Final claim labels induce answer-level signals:
\begin{align*}
\textit{contradiction}(A, C) &\triangleq \exists\, \Clm_i : y_i^* = \textsc{Con}, \\
\textit{baseless}(A, C) &\triangleq \exists\, \Clm_i : y_i^* = \textsc{Nic}.
\end{align*}
We then define the overall hallucination indicator as
\[
\textit{hallucinated}(A, C)
\triangleq
\textit{contradiction}(A, C)
\lor
\textit{baseless}(A, C).
\]

%% file: 03Methodology.tex
\section{Approach}
\label{sec:method}

\subsection{Theoretical Foundation: Retromorphic Testing}
\label{sec:rt-foundation}

We cast context-faithfulness detection as an instance of \emph{retromorphic testing}~\cite{yu2023retromorphic}, a black-box testing approach designed to address the \emph{test oracle problem}.
In many generative settings, including retrieval-augmented generation (RAG), there is no single gold answer for a given input:
multiple responses may be acceptable for the same context--question pair $(C,Q)$.
The relevant criterion is therefore not an exact match to a reference answer, but whether the generated answer is supported by the retrieved context.

Intuitively, retromorphic testing verifies an output by mapping it back to the input domain and checking whether it is consistent with the original inputs.
It couples a forward program $P$ with an auxiliary backward program $B$.
The forward program produces the system output, while the backward program analyzes that output and generates a verification trace that can be checked against the inputs.

In our setting, $P$ represents the RAG generator that maps the context and question $(C,Q)$ to an answer $A$, whereas $B$ acts as a verifier that consumes $(C,Q,A)$ and produces a verification trace $T$:
\begin{equation}
\begin{aligned}
  A &= P(C,Q),\\
  T &= B(C,Q,\, A), \quad \text{check } \mathsf{RR}((C,Q),\, T).
\end{aligned}
\label{eq:retromorphic}
\end{equation}

The retromorphic relation $\mathsf{RR}$ acts as the \emph{test oracle}: it determines whether every element of the trace $T$ is traceable to and supported by the original inputs $(C,Q)$.
The backward program, therefore, produces a structured representation that can be checked directly against the source context.
In RT4CHART, this representation is defined as
\[
T = \{(\Clm_i, y_i^*, E_i^*)\}_{i=1}^{n},
\]
where each tuple contains a claim $\Clm_i$, its final claim label $y_i^*$, and its final evidence spans $E_i^*$ in the context $C$.
The relation $\mathsf{RR}((C,Q), T)$ holds when, for every claim, the extracted evidence $E_i^*$ is grounded in $C$ and justifies the final claim label $y_i^*$.
For entailed claims, $E_i^*$ contains supporting evidence spans.
For contradicted claims, it contains contradicting evidence spans.
For baseless claims, $E_i^* = \varnothing$ (Eq.~\eqref{eq:nic-empty}).

\subsection{Pipeline Overview}

RT4CHART implements the backward program $B$ in four stages (Figure~\ref{fig:overview}):

\begin{figure*}[t]
  \centering
  \includegraphics[width=0.93\textwidth]{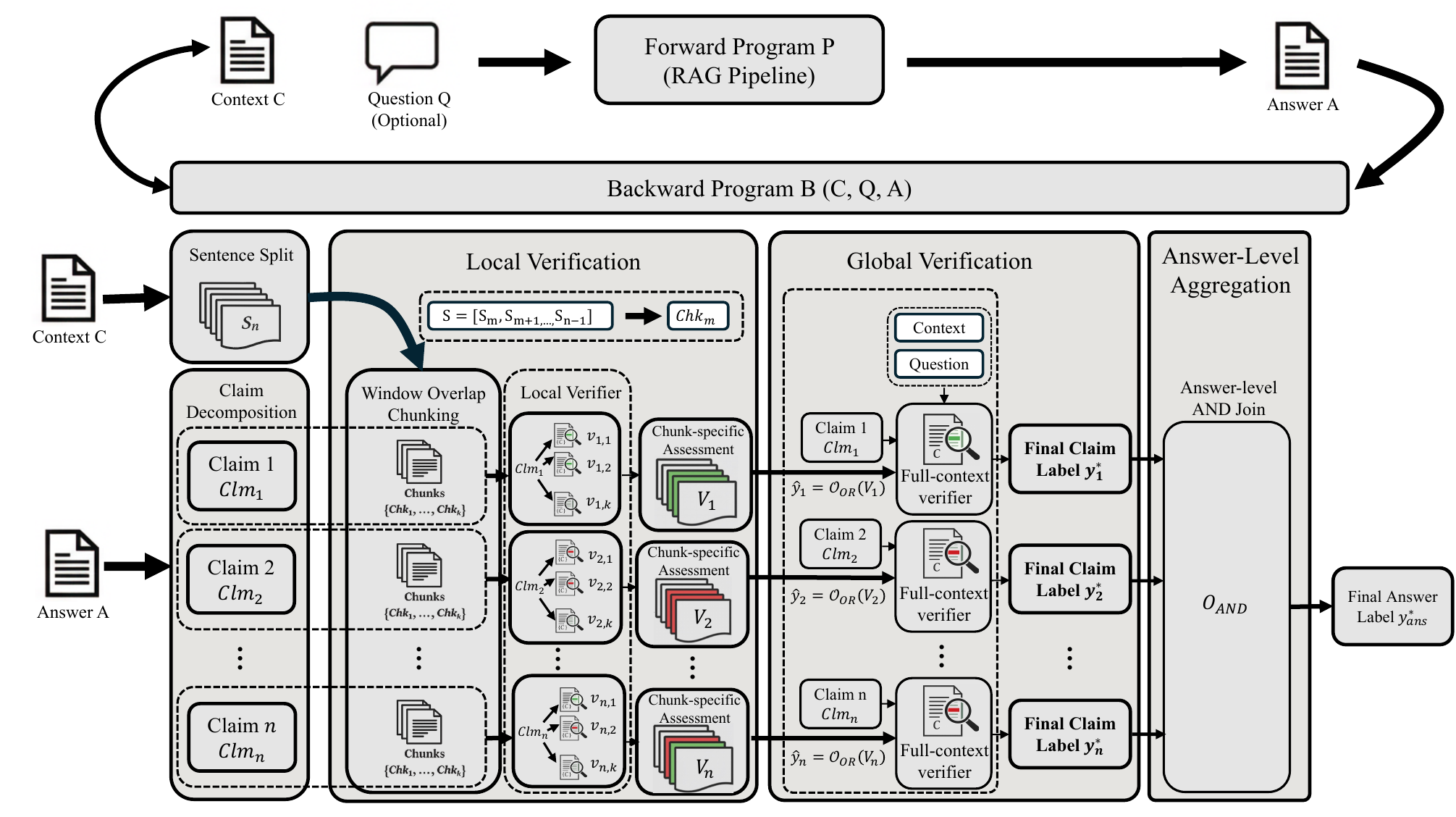}
  \caption{Overview of RT4CHART.
  The system decomposes $A$ into claims, verifies them locally and globally against $C$, and returns claim-level and answer-level outputs.}
  \Description{A pipeline overview of RT4CHART showing claim decomposition, local verification over overlapping context chunks, global verification over the full context, answer-side localization, context-side evidence extraction, and answer-level aggregation.}
  \label{fig:overview}
\end{figure*}

\begin{enumerate}
  \item \textbf{Claim decomposition} (Section~\ref{sec:claim-decomposition}). The answer $A$ is decomposed into self-contained claims that can be verified independently against the retrieved context.
  \item \textbf{Local verification} (Section~\ref{sec:local-verification}). The context is divided into overlapping chunks. For each claim, the verifier scans these chunks for evidence that supports or conflicts with it. The resulting chunk-specific assessments are consolidated into a single \emph{local claim label} via an \textbf{OR-join}.
  \item \textbf{Global verification} (Section~\ref{sec:global-verification}). Each claim is then re-verified against the full context $C$, using the local claim label and an optional focus chunk only as search hints.
  \item \textbf{Answer-level aggregation} (Section~\ref{sec:answer-aggregation}). Final claim labels are aggregated via an \textbf{AND-join}: a single unfaithful claim suffices to flag the entire answer.
\end{enumerate}

We revisit the running example from Figure~\ref{fig:prelim-example} throughout this section.

\subsection{Claim Decomposition}
\label{sec:claim-decomposition}

A \emph{claim} is an atomic, self-contained proposition extracted from the answer $A$ that can be independently verified against the evidence.
Here, \emph{self-contained} means that the claim preserves the minimal semantic content needed for verification, without relying on unresolved pronouns, omitted predicates, or cross-sentence context.
Following the decompose-then-verify paradigm~\cite{min2023factscore}, we adopt a sentence-based decomposition strategy: we first segment $A$ into sentences and then use an LLM to decompose each sentence into claims independently.
The decomposition step preserves qualifiers, including negation, quantities, temporal markers, and modality, since dropping them may change the claim's meaning or verifiability.
The decomposition model takes only the answer $A$ as input, without access to $C$, so that claim formation remains faithful to the answer itself rather than being influenced by the retrieved context.

In the running example (Figure~\ref{fig:prelim-example}), this step yields five claims:
\begin{itemize}
    \item $\Clm_1$: ``the policy applies to all users'';
    \item $\Clm_2$: ``verification is not required'';
    \item $\Clm_3$: ``the free trial lasts 14 days'';
    \item $\Clm_4$: ``verified new users can receive a one-time 7-day extension of the free trial'';
    \item $\Clm_5$: ``the trial includes priority email support''.
\end{itemize}
% In this example, $\Clm_4$ requires combining evidence from non-local sentences, whereas $\Clm_5$ has no support anywhere in the context.
In this example, $\Clm_4$ requires combining evidence from multiple context chunks, whereas $\Clm_5$ has no support anywhere in the context.
Table~\ref{tab:local-assessment} shows how these claims are assessed across local context chunks.

\subsection{Local Verification}
\label{sec:local-verification}

RAG contexts often contain many sentences drawn from one or more retrieved passages.
A verifier that reads the entire context at once may miss subtle contradictions or overlook evidence distributed across it.
Local verification, therefore, scans each claim over overlapping sentence chunks in the context to identify supporting or conflicting signals before global verification.

\paragraph{Chunking.}

Given the ordered sentence sequence $\mathbf{S}$ defined in Section~\ref{sec:prelim}, each chunk $\Chk_k$ is formed by concatenating $W$ consecutive sentences in their original order, and adjacent chunks overlap by $O$ sentences.
The overlap reduces boundary artifacts: it is less likely that two consecutive sentences that jointly support a claim fall into different chunks, leaving neither chunk with sufficient evidence on its own.

\paragraph{Chunk-wise local assessment.}

Given a claim $\Clm_i$ and a context chunk $\Chk_k$, an LLM judge (see Section~\ref{sec:exp-setup}) produces a chunk-specific assessment $v_{i,k} \in \mathcal{Y}$, optionally together with provisional evidence spans restricted to that chunk.
For QA tasks, we provide the user question $Q$ as additional task context; for query-free tasks such as summarization, we set $Q=\varnothing$.
For each claim $\Clm_i$, the judge is applied to every chunk $\Chk_k$, yielding the set $V_i$ of chunk-specific assessments
\[
V_i \;=\; \{v_{i,0},\, v_{i,1},\, \ldots,\, v_{i,K-1}\}.
\]

\paragraph{OR-join ($\mathcal{O}_{or}$).}

\label{eq:oexist}
The chunk-specific assessments in $V_i$ are merged into a single local claim label $\hat{y}_i$.
RT4CHART uses a deterministic priority rule over $\mathcal{Y}=\{\textsc{Ent},\textsc{Nic},\textsc{Con}\}$.
The OR-join returns:
\begin{itemize}
  \item \textsc{Con}\; if any label in $V_i$ is \textsc{Con} \emph{(contradiction dominates)};
  \item \textsc{Ent}\; else if any label in $V_i$ is \textsc{Ent} \emph{(one supporting chunk suffices)};
  \item \textsc{Nic}\; otherwise \emph{(no local chunk provides decisive evidence)}.
\end{itemize}
A local \textsc{Nic} label means only that no individual chunk, when examined in isolation, contains sufficient evidence during local verification; it does not imply that the claim is unsupported in the full context.
Formally,
\begin{equation}
  \hat{y}_i \;=\; \mathcal{O}_{or}(V_i).
  \label{eq:stage1-claim}
\end{equation}

\begin{table}[t]
\caption{Local assessment matrix for the running example with $W{=}2$ and $O{=}1$.
The first three columns show chunk-level assessments, followed by the OR-joined local label $\hat{y}_i$ and the final label $y_i^*$.
Because $\Clm_1$ and $\Clm_2$ are \textsc{Con}, the answer-level AND-join returns \textsc{Con} (Section~\ref{sec:answer-aggregation}).}
  \centering
  \footnotesize
  \setlength{\tabcolsep}{6pt}
  \renewcommand{\arraystretch}{1.12}
  \newcommand{\cEnt}{\colorbox{green!18}{\textsc{Ent}}}
  \newcommand{\cCon}{\colorbox{red!18}{\textsc{Con}}}
  \newcommand{\cNic}{\colorbox{gray!18}{\textsc{Nic}}}
  \begin{tabular}{@{}lccccc@{}}
    \toprule
    & \multicolumn{3}{c}{\textbf{Local verification}} & \textbf{OR-join} & \textbf{Final} \\
    \cmidrule(lr){2-4}\cmidrule(l){5-5}\cmidrule(l){6-6}
    \textbf{Claim} & $\Chk_0$ & $\Chk_1$ & $\Chk_2$ & $\hat{y}_i$ & $y_i^*$ \\
    & {\scriptsize $\{s_0,s_1\}$} & {\scriptsize $\{s_1,s_2\}$} & {\scriptsize $\{s_2,s_3\}$} & & \\
    \midrule
    $\Clm_1$ & \cCon & \cCon & \cNic & \cCon & \cCon \\
    $\Clm_2$ & \cCon & \cCon & \cNic & \cCon & \cCon \\
    $\Clm_3$ & \cNic & \cEnt & \cEnt & \cEnt & \cEnt \\
    $\Clm_4$ & \cNic & \cNic & \cNic & \cNic & \cEnt \\
    $\Clm_5$ & \cNic & \cNic & \cNic & \cNic & \cNic \\
    \midrule
    \multicolumn{5}{@{}r}{\textit{Answer-level} (AND-join):} & \cCon \\
    \bottomrule
  \end{tabular}

  \Description{A matrix showing five claims as rows and three overlapping local chunks as columns. Claim texts are omitted for compactness. The first three columns show chunk-specific assessments, followed by the locally aggregated label and the final claim label after global verification. Claim 4 is locally baseless in all chunks, but becomes entailed after global verification. Because Claims 1 and 2 are contradicted and contradiction has priority in the AND-join, the final answer-level label is contradicted.}
  \label{tab:local-assessment}
\end{table}

Continuing the running example in Figure~\ref{fig:prelim-example}, Table~\ref{tab:local-assessment} summarizes local assessment under $W{=}2$ and $O{=}1$, yielding three overlapping chunks.
The first three columns list chunk-level judgments, followed by the OR-joined local label $\hat{y}_i$ and the final label $y_i^*$ after global verification:
$\Chk_0{=}\{s_0,s_1\}$,
$\Chk_1{=}\{s_1,s_2\}$,
and $\Chk_2{=}\{s_2,s_3\}$.
For $\Clm_3$, at least one local chunk contains sufficient support, so the claim is labeled \textsc{Ent}.
For $\Clm_4$, $\Chk_0$ contains $s_0$ and $s_1$ but not $s_3$, while $\Chk_2$ contains $s_3$ but neither $s_0$ nor $s_1$.
No local chunk contains the complete evidence, so the OR-joined local label is \textsc{Nic}; the claim is revised only after the global judge combines $s_0$, $s_1$, and $s_3$, yielding the final label \textsc{Ent}.
$\Clm_5$ remains \textsc{Nic} in every chunk because the context contains no supporting evidence for it.

\subsection{Global Verification}
\label{sec:global-verification}

Local verification examines individual chunks in isolation.
However, this fragmented view can lead to errors: evidence may be distributed across multiple chunks, or an apparent local contradiction may be resolved by surrounding text elsewhere in the context.
Global verification addresses these limitations by re-evaluating each claim against the entire context $C$.

Global verification uses local verification outputs only as search hints.
That is, local verification may suggest which chunk is worth inspecting, but it does not provide evidence that can be directly reused as the final justification.
Instead, the global judge must read the full context $C$ and re-extract the final evidence spans from $C$ itself before assigning the final claim label $y_i^*$.

\paragraph{Adaptive prompting strategy.}
Because different local claim labels (\textsc{Nic}, \textsc{Con}, \textsc{Ent}) potentially lead to different likely failure modes during local verification, we adapt the prompt for global verification accordingly (summarized in Table~\ref{tab:stage2-strategies}):
\begin{itemize}
    \item \textbf{Handling locally \textsc{Nic}-labeled claims:} The claim may still be true, but its evidence may be distributed across chunk boundaries. To catch this case, the judge receives the full context and searches from scratch, with no location hint.
    \item \textbf{Handling locally entailed (\textsc{Ent}) or contradicted (\textsc{Con}) claims:} Local verification found a strong signal in a specific chunk. We pass that chunk to the global judge as a ``location hint.'' The judge then examines this hint in its full context to determine whether the local decision still holds or whether the surrounding text alters its interpretation.
\end{itemize}

For instance, returning to our running example (Figure~\ref{fig:prelim-example}), $\Clm_4$ is locally \textsc{Nic} because $\Chk_0$ contains $s_0$ and $s_1$ but not $s_3$, whereas $\Chk_2$ contains $s_3$ but neither $s_0$ nor $s_1$.
The global judge combines $s_0$ (free-trial scope), $s_1$ (eligibility limited to identity-verified new users), and $s_3$ (one seven-day extension for eligible users) and assigns $\Clm_4$ the final label \textsc{Ent}.
By contrast, $\Clm_5$ (``priority email support'') remains \textsc{Nic} even after global verification, confirming that it is genuinely baseless.

\begin{table}[t]
\centering
\scriptsize
\setlength{\tabcolsep}{3pt}
\caption{Global verification strategy conditioned on the local claim label.}
\label{tab:stage2-strategies}
\begin{tabular}{@{}p{0.14\columnwidth}p{0.23\columnwidth}p{0.23\columnwidth}p{0.32\columnwidth}@{}}
\toprule
Local label & Prompt input & Likely local failure mode & Verification goal \\
\midrule
\textsc{Nic} & Full $C$ + claim &
No single chunk contains enough evidence &
Confirm baseless or revise to entailed \\
\textsc{Con} & Full $C$ + focus chunk + claim &
Local contradiction, globally resolved &
Confirm contradiction or revise \\
\textsc{Ent} & Full $C$ + focus chunk + claim &
Locally entailed, globally overridden &
Confirm entailment or revise \\
\bottomrule
\end{tabular}
\end{table}

\subsection{Answer-Level Aggregation}
\label{sec:answer-aggregation}

\paragraph{AND-join ($\mathcal{O}_{and}$).}
\label{eq:osafe}
After global verification assigns a final claim label $y_i^*$ to each claim, these labels are aggregated into an answer-level verdict.
RT4CHART uses a deterministic AND-join.
Given a set of labels $Z \subseteq \mathcal{Y}$, the AND-join returns:
\begin{itemize}
  \item \textsc{Con}\; if any label in $Z$ is \textsc{Con} \emph{(contradiction dominates)};
  \item \textsc{Nic}\; else if any label in $Z$ is \textsc{Nic} \emph{(missing evidence blocks full entailment)};
  \item \textsc{Ent}\; otherwise \emph{(all claims are satisfied)}.
\end{itemize}
This aggregation is intentionally strict: a single unfaithful claim is sufficient to flag the entire answer.
Formally, the answer-level label is
\begin{equation}
  y_{\text{ans}}^* \;=\; \mathcal{O}_{and}(\{y_1^*,\ldots,y_n^*\}).
  \label{eq:final-answer}
\end{equation}
This design ensures that a partially hallucinated answer is still flagged rather than masked by faithful claims.

%% file: 04Experiments.tex
\section{Experiments}
\label{sec:experiments}

This section evaluates RT4CHART as a \emph{context-faithfulness} detector, operating under the strict assumption that only the provided context serves as evidence.
The experiments are structured around four research questions (RQs).

\subsection{Research Questions}
\label{sec:rqs}

\noindent\textbf{RQ1. Overall effectiveness:} Under the same $(C,Q,A)$ protocol, how effective is RT4CHART on answer-level hallucination detection and span-level localization?

\noindent\textbf{RQ2. Contribution of hierarchical verification:} How do local verification and global verification each contribute to answer-level detection and span-level localization?

\noindent\textbf{RQ3. Design sensitivity:} How sensitive is RT4CHART to claim decomposition granularity and chunking hyperparameters?

\noindent\textbf{RQ4. Robustness and cost:} How robust is RT4CHART across judge models and repeated runs, and what cost does it incur?

\subsection{Experimental Setup}
\label{sec:exp-setup}

\paragraph{Datasets.}
We evaluate on two benchmarks.
\textbf{RAGTruth++} is a re-annotation of a 408-example QA/summarization subset of the RAGTruth test set~\cite{niu2024ragtruth}, with substantially more complete hallucination-span annotations than the original subset annotations.
\textbf{RAGTruth-Enhance} is our broader re-annotation of the RAGTruth evaluation split.
Unlike RAGTruth++, RAGTruth-Enhance covers the full RAGTruth evaluation split, including QA, summarization, and data-to-text tasks.

\paragraph{Annotation completeness and the need for RAGTruth-Enhance.}
\label{sec:ragtruth-enhance}
RAGTruth~\cite{niu2024ragtruth} is a popular benchmark for RAG hallucination detection, but its original span annotations contain substantial false negatives.
On the 408-example QA/summarization subset later released as \textbf{RAGTruth++}, re-annotation increased the number of annotated hallucination spans from 86 to 865, indicating that the original labels substantially under-count hallucination spans~\cite{ragtruthpp_blog2025,blueguardrails2025ragtruthpp}.
This motivates a broader re-annotation of the RAGTruth evaluation split, which we release as \textbf{RAGTruth-Enhance}.

\paragraph{Construction of RAGTruth-Enhance.}
RAGTruth-Enhance is built with a disagreement-driven two-stage protocol:
\begin{enumerate}
  \item We first use an LLM-based auditor to flag samples whose judgments differ from the original annotation, while leaving matched cases unchanged.
  As a sanity check, we randomly inspect \resRagTruthEnhAgreeSanityN\ matched cases and observe accuracy above \resRagTruthEnhAgreeSanityAgreement, supporting the treatment of these cases as low priority for further review.

  \item For the flagged subset (\resRagTruthEnhConflictN\ cases, approximately \resRagTruthEnhDisagreePct\ of the data), two authors independently review each case and resolve disagreements through discussion.
  Residual disputes are rare ($<\resRagTruthEnhResidualDisputePct$ of the reviewed cases).
\end{enumerate}
Analogous to the answer-side hallucination spans, RAGTruth-Enhance also includes refuting context evidence for contradiction labels, which enables the context-evidence grounding evaluation reported later.

\paragraph{Compared detectors.}
% We compare \textbf{RT4CHART} with \textbf{Lettuce}~\cite{kovacs2025lettucedetecthallucinationdetectionframework},
% \textbf{MetaQA}~\cite{yang2025metaqa},
% \textbf{SelfCheckGPT}~\cite{manakul2023selfcheckgpt},
% and \textbf{Vectara}.
% RT4CHART and Lettuce produce localized span outputs, whereas MetaQA, SelfCheckGPT, and Vectara are answer-level baselines.
% Lettuce is a supervised token-level span tagger based on ModernBERT, while Vectara is a commercial factual-consistency API.

We compare \textbf{RT4CHART} against a diverse set of strong baselines chosen to cover both \emph{span-level} and \emph{answer-level} hallucination detection, as well as both \emph{open-source} and \emph{closed-source} systems.
Specifically, we include \textbf{Lettuce}~\cite{kovacs2025lettucedetecthallucinationdetectionframework} as a strong open-source span-level detector, \textbf{MetaQA}~\cite{yang2025metaqa} and \textbf{SelfCheckGPT}~\cite{manakul2023selfcheckgpt} as representative open-source \emph{LLM-driven} answer-level baselines, and \textbf{Vectara} as a strong closed-source commercial factual-consistency system.
RT4CHART and Lettuce produce localized span outputs, whereas MetaQA, SelfCheckGPT, and Vectara are answer-level baselines.
Lettuce is a supervised token-level span tagger based on ModernBERT, while Vectara is a commercial factual-consistency API.

\paragraph{Unified evaluation contract.}
All methods are evaluated under the same $(C,Q,A)$ contract, where the retrieved context $C$, user query $Q$, and model answer $A$ are provided as input.
Their outputs are mapped to the common label space $\{\textsc{Ent}, \textsc{Con}, \textsc{Nic}\}$ defined in Section~\ref{sec:prelim}.
For methods that do not return span-level predictions, span-level evaluation is not applicable and is therefore omitted.

\paragraph{Context-only adaptation of non-native baselines.}
\label{sec:ctx-adaptation}
MetaQA, SelfCheckGPT, and Vectara were not originally designed for our strict retrieved-context setting, in which $C$ is the only admissible evidence source.
We therefore adapt them with minimal changes.
Specifically, we (1) use $C$ as the sole evidence source, (2) retain each baseline's original inference workflow and decision procedure as much as possible, and (3) map outputs into the common label space.
This design keeps the comparison as faithful as possible to the original methods while enforcing the same context-only evaluation setting for all detectors.

\paragraph{Baseline-specific implementations.}
\begin{itemize}
  \item \textbf{MetaQA (context-adapted).}
    We adapt MetaQA to the retrieved-context setting by reusing its mutation-generation prompts on the $(Q,A)$ pair and replacing open-world verification with verification against the provided context $C$.
    We aggregate mutation-based verification signals into an answer-level prediction rather than producing span-level outputs.

  \item \textbf{SelfCheckGPT (context-adapted).}
  We follow the upstream LLM-prompt baseline: sample $n{=}10$ alternative answers for the same $(C,Q)$, query whether $A$ is supported by each sampled passage, map the resulting Yes/No/N/A votes to $0.0/1.0/0.5$, and average them into a scalar \emph{inconsistency} score.
  Using a threshold sweep on our validation runs, we found $0.5$ to be a more stable threshold in our setting, and therefore use it to map the continuous score to \textsc{Ent} versus \textsc{Nic}.\footnote{Although the prompt asks about support, the upstream prompt baseline encodes \textsc{Yes}$\to 0$, \textsc{No}$\to 1$, so lower scores indicate stronger support and higher scores indicate greater unsupportedness / hallucination risk. The baseline does not explicitly distinguish contradiction from missing support.}

  \item \textbf{Vectara.}
  We use Vectara's HHEM~\cite{vectara_hallucination_eval_docs,hhem-2.1-open}, which returns an answer-level factual-consistency score for a generated answer given source passages.
  We pass the full answer as the hypothesis and the full provided context $C$ in a single API call.
  Following Vectara's recommended threshold, we use $0.5$: scores below this threshold are mapped to \textsc{Con}, and \textsc{Nic} is not predicted.
\end{itemize}

\paragraph{Implementation details.}
Unless otherwise specified, RT4CHART employs judge model \resJudgeModel, claim decomposition setting \resExtractorMode, chunking window/overlap $(W,O)=(\resChunkWindow,\resChunkOverlap)$, and global verification (Section~\ref{sec:global-verification}).
We set the random seed to \resRandomSeed; claim decomposition, local verification, and global verification all use temperature $\resStageTwoTemperature$.

% \paragraph{Metrics.}
% For answer-level evaluation, we report precision, recall, and F1 on the hallucination indicator
% $\textit{hallucinated}(A,C)$ defined in Section~\ref{sec:prelim}, where an answer is positive if any claim is labeled \textsc{Con} or \textsc{Nic}.
% We use F1 as the primary summary metric for comparability with prior work, while also reporting precision and recall because missed hallucinations are especially costly in auditing settings.

% overlaps
\paragraph{Metrics.}
We evaluate all methods on:

\textbf{Answer-level hallucination detection.}
    We report precision, recall, and F1 on the hallucination indicator defined in Section~\ref{sec:prelim}. An answer is positive if any claim is labeled \textsc{Con} or \textsc{Nic}; we use F1 as the primary summary metric for overall comparison.

\textbf{Answer-side span localization.}
    We follow the overlap-based protocol of RAGTruth~\cite{niu2024ragtruth} and evaluate the overlap between the predicted and gold hallucinated text spans in the original answer. For each sample $s$, let $G_s$ and $P_s$ denote the union of gold and predicted hallucinated character offsets in the answer. We compute
    \[
    \mathrm{TP} = \sum_s |G_s \cap P_s|,\qquad
    \mathrm{FP} = \sum_s |P_s \setminus G_s|,\qquad
    \mathrm{FN} = \sum_s |G_s \setminus P_s|,
    \]
    and report micro precision, recall, and F1.

\textbf{Context-side evidence grounding.}
    For contradiction cases with mappable gold refuting evidence, we additionally evaluate context-side evidence grounding using the same overlap-based scoring. This complementary diagnostic measures whether correctly detected contradictions are supported by localized refuting evidence and is reported on RAGTruth-Enhance for systems that return such spans.

\subsection{Overall Effectiveness}
\label{sec:main-results}

\textbf{RQ1 investigates whether RT4CHART improves both answer-level hallucination detection and span-level localization under the same $(C,Q,A)$ evaluation contract.}
We answer this question using the answer-level results in Table~\ref{tab:main_answer_results} and the span-level localization results in Table~\ref{tab:localization_diagnostics}.

\begin{table}[t]
  \caption{Main answer-level hallucination detection results on RAGTruth++ and RAGTruth-Enhance (Precision/Recall/F1).
  Methods marked with ``*'' are adapted to the context-only setting.}
  \centering
  \small
  \setlength{\tabcolsep}{6pt}
  \begin{tabular}{lccc}
    \toprule
    Detector & Precision & Recall & F1 \\
    \midrule
    \multicolumn{4}{l}{\textbf{RAGTruth++}} \\
    \resRagTruthppMainRowRTFourCHART \\
    \resRagTruthppMainRowLettuce \\
    \resRagTruthppMainRowMetaQA \\
    \resRagTruthppMainRowSelfCheck \\
    \resRagTruthppMainRowVectara \\
    \midrule
    \multicolumn{4}{l}{\textbf{RAGTruth-Enhance}} \\
    \resRagTruthEnhMainRowRTFourCHART \\
    \resRagTruthEnhMainRowLettuce \\
    \resRagTruthEnhMainRowMetaQA \\
    \resRagTruthEnhMainRowSelfCheck \\
    \resRagTruthEnhMainRowVectara \\
    \bottomrule
  \end{tabular}
  \label{tab:main_answer_results}
\end{table}

\begin{table}[t]
  \caption{Answer-side span localization results on RAGTruth-Enhance and RAGTruth++ (Precision/Recall/F1).} % ts.}
  \centering
  \small
  \setlength{\tabcolsep}{6pt}
  \begin{tabular}{llccc}
    \toprule
    Dataset & Detector & Precision & Recall & F1 \\
    \midrule
    RAGTruth-Enhance & RT4CHART & \resRagTruthEnhLocPrec & \textbf{\resRagTruthEnhLocRec} & \textbf{\resRagTruthEnhLocFOne} \\
    RAGTruth-Enhance & Lettuce & \textbf{\resRagTruthEnhLocPrecLettuce} & \resRagTruthEnhLocRecLettuce & \resRagTruthEnhLocFOneLettuce \\
    RAGTruth++ & RT4CHART & \resRagTruthppLocAnyOverlapPrec & \textbf{\resRagTruthppLocAnyOverlapRec} & \textbf{\resRagTruthppLocAnyOverlapFOne} \\
    RAGTruth++ & Lettuce & \textbf{\resRagTruthppLocAnyOverlapPrecLettuce} & \resRagTruthppLocAnyOverlapRecLettuce & \resRagTruthppLocAnyOverlapFOneLettuce \\
    \bottomrule
  \end{tabular}
  \label{tab:localization_diagnostics}
\end{table}
 
\paragraph{Main results.}
Across both benchmarks, RT4CHART achieves the highest answer-level F1 among all compared methods (Table~\ref{tab:main_answer_results}).
Its advantage stems primarily from stronger recall, particularly on the larger, more diverse RAGTruth-Enhance benchmark, suggesting that claim-based local processing helps detect partially hallucinated answers that holistic answer scorers often miss.
This recall-oriented behavior is particularly desirable in practical auditing settings, where missing a hallucination can be substantially more costly than flagging an additional suspicious case.
Among the baselines, Vectara is the strongest answer-level competitor.
On RAGTruth++ and RAGTruth-Enhance, RT4CHART improves over the best baseline by $\Delta$F1$\,{=}\,{+}\resRagTruthppAnyFOneDeltaBestBaseline$ and $+\resRagTruthEnhAnyFOneDeltaBestBaseline$, respectively.

\paragraph{Span-level localization.}
Under this span-level evaluation rule, RT4CHART achieves the highest localization F1 on both datasets (Table~\ref{tab:localization_diagnostics}).
Compared with Lettuce, RT4CHART attains substantially higher recall, while Lettuce achieves higher precision.
A likely reason is that RT4CHART localizes hallucinations at a coarser, often sentence-level granularity, which tends to include additional non-hallucinated text, thereby lowering precision under overlap-based scoring.

\paragraph{Context-evidence diagnostic on contradiction cases.}

While answer-side span localization (Table~\ref{tab:localization_diagnostics}) captures whether the system can localize unsupported text in the \emph{answer}, we further ask whether correct contradiction predictions are also grounded in the right \emph{refuting context evidence}. For this diagnostic only, we additionally include a \textbf{GPT-4o mini (direct)} baseline that uses a single direct prompt over the full context and answer to produce an answer-level judgment in the common label space, along with answer-span and refuting-evidence spans. RAGTruth-Enhance includes refuting context evidence for contradiction labels, enabling this analysis on the contradiction subset. We map each gold evidence string back to sentence-local offsets in the source context. We evaluate contradiction detection at the answer level. For grounding, we compute overlap-based precision, recall, and F1 only on correctly detected contradiction cases with alignable gold evidence. Refuting-evidence grounding is reported as a conditional diagnostic on each system's correctly detected contradiction cases.

\begin{table}[t]
  \caption[Contradiction-case detection and refuting-evidence grounding on RAGTruth-Enhance]{Contradiction-case detection and refuting-evidence grounding on RAGTruth-Enhance (Precision/Recall/F1).}
  \centering
  \small
  \setlength{\tabcolsep}{4pt}
  \begin{tabular}{@{}p{0.25\columnwidth}p{0.25\columnwidth}ccc@{}}
    \toprule
    Task & Method & Precision & Recall & F1 \\
    \midrule
    \multirow{2}{=}{\centering\shortstack{Contradiction-case\\detection}}
      & RT4CHART & \resRagTruthEnhConflictCasePrec & \resRagTruthEnhConflictCaseRec & \resRagTruthEnhConflictCaseFOne \\
      & \shortstack[l]{GPT-4o mini\\(direct)} & \resRagTruthEnhConflictCasePrecGptMini & \resRagTruthEnhConflictCaseRecGptMini & \resRagTruthEnhConflictCaseFOneGptMini \\
    \midrule
    \multirow{2}{=}{\centering\shortstack{Refuting-evidence\\grounding}}
      & RT4CHART & \resRagTruthEnhConflictEvidenceSpanPrec & \resRagTruthEnhConflictEvidenceSpanRec & \resRagTruthEnhConflictEvidenceSpanFOne \\
      & \shortstack[l]{GPT-4o mini\\(direct)} & \resRagTruthEnhConflictEvidenceSpanPrecGptMini & \resRagTruthEnhConflictEvidenceSpanRecGptMini & \resRagTruthEnhConflictEvidenceSpanFOneGptMini \\
    \bottomrule
  \end{tabular}
  \label{tab:context_evidence_conflict}
\end{table}

As shown in Table~\ref{tab:context_evidence_conflict}, RT4CHART achieves higher contradiction-case detection F1 than the direct GPT-4o mini baseline (\resRagTruthEnhConflictCaseFOne\ vs.\ \resRagTruthEnhConflictCaseFOneGptMini). The conditional grounding diagnostic evaluates whether a correctly detected contradiction is supported by localized refuting evidence; RT4CHART obtains \resRagTruthEnhConflictEvidenceSpanFOne\ F1 on this diagnostic.

\begin{rqbox}
\textbf{Answer to RQ1:} RT4CHART performs best overall: it achieves the highest answer-level F1 and span-level localization F1 on both RAGTruth++ and RAGTruth-Enhance. Its gains come primarily from higher recall. For contradiction cases, it reports conditional refuting-evidence grounding results, with significant room for improvement.
\end{rqbox}

% \subsection{Contribution of Hierarchical Verification}
\subsection{Complementary Roles of Local and Global Verification}
\label{sec:stage2-ablation}

\textbf{RQ2 asks how hierarchical verification contributes to performance.}
We examine this question from two sides: by ablating global verification, and by testing a direct variant that bypasses claim decomposition and local verification.

\begin{table}[t]
  \caption{Effect of global verification on answer-level hallucination detection (Precision/Recall/F1).}
  \centering
  \small
  \begin{tabular}{@{}llccc@{}}
    \toprule
    Dataset & Setting & Precision & Recall & F1 \\
    \midrule
    \multirow{2}{*}{RAGTruth++}
      & w/ global verif. & \resRagTruthppStageTwoWithAnyPrec & \textbf{\resRagTruthppStageTwoWithAnyRec} & \textbf{\resRagTruthppStageTwoWithAnyFOne} \\
      & w/o global verif. & \textbf{\resRagTruthppStageTwoNoAnyPrec} & \resRagTruthppStageTwoNoAnyRec & \resRagTruthppStageTwoNoAnyFOne \\
    \midrule
    \multirow{2}{*}{RAGTruth-Enhance}
      & w/ global verif. & \resRagTruthEnhStageTwoWithAnyPrec & \textbf{\resRagTruthEnhStageTwoWithAnyRec} & \textbf{\resRagTruthEnhStageTwoWithAnyFOne} \\
      & w/o global verif. & \textbf{\resRagTruthEnhStageTwoNoAnyPrec} & \resRagTruthEnhStageTwoNoAnyRec & \textbf{\resRagTruthEnhStageTwoNoAnyFOne} \\
    \bottomrule
  \end{tabular}
  \label{tab:stage2_ablation}
\end{table}

\paragraph{Effect of global verification.}
Table~\ref{tab:stage2_ablation} reports the comparison.
On RAGTruth++, global verification increases recall from 0.655 to 0.718 while leaving precision nearly unchanged (0.851 vs.\ 0.845), yielding an F1 gain of 0.036.
On RAGTruth-Enhance, however, global verification does not improve overall answer-level performance ($\Delta$F1$\,{=}\,$\resRagTruthEnhStageTwoDeltaAnyFone): it yields a large gain on baseless detection ($\Delta$F1$\,{=}\,$\resRagTruthEnhStageTwoDeltaBaselessFone) but no corresponding overall improvement.
This pattern shows that global verification is complementary rather than uniformly beneficial. Its clearest overall benefit appears on the more challenging RAGTruth++ benchmark, where supporting evidence may be distributed across multiple chunks and cannot be recovered solely from chunk-local verification.

\paragraph{Contribution of claim-based local processing.}
We next examine the reverse setting: a variant that bypasses claim decomposition and local verification entirely, and instead predicts hallucination spans directly from $(C,Q,A)$ with a single prompt.
Table~\ref{tab:gptspan_direct_baseline} reports the resulting performance on RAGTruth++.
Its span-level localization performance is substantially worse than that of the full pipeline, trailing RT4CHART by \resDirectSpanMiniSpanGapVsRT~points in span F1, showing that claim-based local processing provides the primary observed gain rather than an optional refinement.

\begin{table}[t]
  \caption{RT4CHART versus the direct span-prediction variant on RAGTruth++ (answer-level Precision/Recall/F1 and span F1).}
  \centering
  \small
  \begin{tabular}{@{}lcccc@{}}
    \toprule
    Variant & Precision & Recall & F1 & Span F1 \\
    \midrule
    RT4CHART (full pipeline) & \resRagTruthppFullPipelineAnswerPrec & \textbf{\resRagTruthppFullPipelineAnswerRec} & \textbf{\resRagTruthppFullPipelineAnswerFOne} & \textbf{\resRagTruthppFullPipelineSpanMicroFOne} \\
    \resDirectSpanModelMini\ (direct spans) & \textbf{\resDirectSpanMiniAnswerPrec} & \resDirectSpanMiniAnswerRec & \resDirectSpanMiniAnswerFOne & \resDirectSpanMiniSpanMicroFOne \\
    \bottomrule
  \end{tabular}
  \label{tab:gptspan_direct_baseline}
\end{table}

\begin{rqbox}
\textbf{Answer to RQ2:} Claim-based local processing provides the primary observed gain. Decomposing the answer into verifiable units and aligning them with localized evidence enables fine-grained hallucination localization that the direct variant cannot match. Global verification is complementary: it is most useful when supporting evidence is distributed across multiple chunks, but its benefit is selective rather than uniform across datasets.
\end{rqbox}

\subsection{Design Sensitivity}
\label{sec:ablations}

{\textbf{RQ3 examines how sensitive RT4CHART is to key design choices.}
We focus on two factors that shape the verification process: the granularity of claim decomposition and the hyperparameters used for localized evidence screening.
Unless otherwise noted, all analyses in this section are conducted on RAGTruth-Enhance.}

\paragraph{Sensitivity to claim decomposition granularity.}
Table~\ref{tab:ablations} compares two decomposition strategies: \textbf{Sentence-based}, which extracts claims sentence by sentence, and \textbf{Holistic}, which reads the full answer once and then produces a claim set for verification.
The two strategies achieve similar overall answer-level F1 (\resAblateExtractorFocusAnyFOne\ vs.\ \resAblateExtractorOnceAnyFOne), but exhibit a consistent precision--recall trade-off across task types.
Sentence-based decomposition yields higher recall (e.g., \resAblateExtractorFocusQARec\ on QA), whereas Holistic decomposition tends to achieve higher precision (e.g., \resAblateExtractorOnceQAPrec\ on QA).
The same pattern holds for Summary and Data2Text.
These results suggest that RT4CHART is robust to decomposition style, although granularity affects error preference: Sentence-based decomposition is more likely to surface localized or partial hallucinations, while Holistic decomposition is more conservative.

The two strategies also differ in \emph{span-level traceability}.
Sentence-based decomposition extracts claims from each sentence individually, so every claim retains a direct link to its source sentence; when a claim is judged unfaithful, the verdict traces back to the originating sentence span in the answer (Table~\ref{tab:localization_diagnostics}).
Holistic decomposition extracts claims from the full answer at once and does not track which sentence each claim originates from, so it cannot produce span-level output.
This structural advantage is the main reason RT4CHART adopts Sentence-based decomposition.

\begin{table}[t]
  \caption{Decomposition ablation on RAGTruth-Enhance.
  Sentence-based and Holistic yield similar overall F1, with a precision--recall trade-off by task type.}
  \centering
  \small
  \begin{tabular}{@{}llccc@{}}
    \toprule
    Task & Mode & Precision & Recall & F1 \\
    \midrule
    \multirow{2}{*}{QA}
      & Sent. & \resAblateExtractorFocusQAPrec & \textbf{\resAblateExtractorFocusQARec} & \textbf{\resAblateExtractorFocusQAFOne} \\
      & Hol.  & \textbf{\resAblateExtractorOnceQAPrec} & \resAblateExtractorOnceQARec & \resAblateExtractorOnceQAFOne \\
    \midrule
    \multirow{2}{*}{Summary}
      & Sent. & \resAblateExtractorFocusSummPrec & \textbf{\resAblateExtractorFocusSummRec} & \textbf{\resAblateExtractorFocusSummFOne} \\
      & Hol.  & \textbf{\resAblateExtractorOnceSummPrec} & \resAblateExtractorOnceSummRec & \resAblateExtractorOnceSummFOne \\
    \midrule
    \multirow{2}{*}{Data2Text}
      & Sent. & \resAblateExtractorFocusDTTPrec & \textbf{\resAblateExtractorFocusDTTRec} & \textbf{\resAblateExtractorFocusDTTFOne} \\
      & Hol.  & \textbf{\resAblateExtractorOnceDTTPrec} & \resAblateExtractorOnceDTTRec & \resAblateExtractorOnceDTTFOne \\
    \midrule
    \multirow{2}{*}{Overall}
      & Sent. & -- & -- & \textbf{\resAblateExtractorFocusAnyFOne} \\
      & Hol.  & -- & -- & \resAblateExtractorOnceAnyFOne \\
    \bottomrule
  \end{tabular}
  \label{tab:ablations}
\end{table}

\paragraph{Sensitivity to chunking hyperparameters.}
We next vary the chunk window and overlap parameters $(W,O)$ over a small grid and report the resulting answer-level performance in Table~\ref{tab:rq3-chunking-grid}.
The main pattern is stability rather than sharp sensitivity.
Once the window is large enough to preserve local evidence dependencies ($W \ge \resChunkAdequateWindow$), performance remains strong across nearby settings.
The best configuration is $(\resChunkBestWindow,\resChunkBestOverlap)$ with F1$=\resChunkBestFOne$, and among the high-coverage settings, the spread is at most \resChunkHighCoverageDeltaFOne.
This suggests that RT4CHART does not rely on narrowly tuned chunk boundaries, provided the local window is sufficiently wide to retain the evidence needed for local verification.

\begin{table}[t]
  \caption{Chunking sensitivity on RAGTruth-Enhance under different $(W,O)$ settings (answer-level P/R/F1).}
  \centering
  \small
  \begin{tabular}{lcc}
    \toprule
    $(W,O)$ & P/R/F1 \\
    \midrule
    \resChunkGridRowFifteenFive \\
    \resChunkGridRowFifteenTen \\
    \resChunkGridRowTwentyFiveFive \\
    \resChunkGridRowTwentyFiveTen \\
    \resChunkGridRowThirtyFiveFive \\
    \resChunkGridRowThirtyFiveTen \\
    \bottomrule
  \end{tabular}
  \label{tab:rq3-chunking-grid}
\end{table}

% \paragraph{Overall interpretation.}
% Overall, RT4CHART is configurable but not fragile.
% Sentence-based decomposition outperforms Holistic decomposition: it achieves higher recall and comparable F1 while uniquely supporting span-level localization.
% Chunking performance is stable once the chunk window is sufficiently large to preserve local evidence dependencies.

\begin{rqbox}
\textbf{Answer to RQ3:} RT4CHART is configurable without being brittle. Sentence-based decomposition is the better default because it preserves comparable F1, improves recall, and uniquely enables span-level localization, while chunking remains stable once the local window is large enough to retain the evidence needed for verification.
\end{rqbox}

\subsection{Robustness, Reproducibility, and Cost}
\label{sec:robustness}

{\textbf{RQ4 asks whether RT4CHART is reliable under model variation and repeated runs, and what cost it incurs in practice.}
We examine this question from three perspectives: robustness to the choice of decomposition model and judge model, run-to-run reproducibility, and API cost under realistic evaluation workloads.}

\paragraph{Robustness to judge-model choice.}
As a stronger, more expensive alternative to the default judge, we replace \resJudgeModel\ with \resAltJudgeModel, an open-weight, MIT-licensed alternative.
This substitution tests whether RT4CHART's verification behavior transfers across LLM backends rather than depending on a single low-cost judge.
We rerun the full RT4CHART pipeline---claim decomposition, local verification, and global verification---under the same context-only protocol and hyperparameter settings.
Table~\ref{tab:rt4chart_glm_stability} reports the results.
Across both benchmarks, performance remains broadly stable under this model swap, suggesting that RT4CHART does not rely on a specific judge model to function effectively.

\begin{table}[t]
  \caption{Cross-model robustness when swapping the decomposition model and judge model (answer-level Precision/Recall/F1).}
  \centering
  \small
  \setlength{\tabcolsep}{6pt}
  \begin{tabular}{llccc}
    \toprule
    Dataset & Model & Precision & Recall & F1 \\
    \midrule
    \resRagTruthppRTFourCHARTJudgePRFRow \\
    \resRagTruthppRTFourCHARTAltPRFRow \\
    \midrule
    \resRagTruthEnhRTFourCHARTJudgePRFRow \\
    \resRagTruthEnhRTFourCHARTAltPRFRow \\
    \bottomrule
  \end{tabular}
  \label{tab:rt4chart_glm_stability}
\end{table}

\paragraph{Run-to-run reproducibility.}
We next examine whether the pipeline remains stable across repeated executions.
Because LLM serving APIs do not guarantee strict determinism, identical inputs can still produce slightly different outputs across calls.
To quantify this inherent variance, we execute RT4CHART \resStabilityRuns\ times on RAGTruth++ under the same configuration (temperature $\resStageTwoTemperature$ and all other hyperparameters unchanged) and measure the spread of the resulting metrics.
Across runs, RT4CHART achieves a mean answer-level F1 of $\resStabilityAnswerMean \pm \resStabilityAnswerStd$ and a span-level localization F1 of $\resStabilitySpanMean \pm \resStabilitySpanStd$.
These small standard deviations indicate that the reported gains are not artifacts of stochastic instability, but remain consistent across repeated executions.

\paragraph{Deployment cost under the default judge model.}
\label{sec:cost_analysis}
Finally, we analyze the computational cost of RT4CHART to assess whether its verification procedure is practical for routine context-faithfulness auditing.
Using \resJudgeModel\ on RAGTruth-Enhance ($N=\resRagTruthEnhanceN$) corresponds to a total API cost of \$\resCostTotalUsd\ USD, or approximately \textbf{\$\resCostPerSampleUsd\ per sample}.
% Given that RT4CHART performs claim-level decomposition, localized evidence matching, and structured verification, this per-sample cost is low enough to support frequent regression testing in practice.
% In particular, it enables teams to re-audit the context-faithfulness of RAG outputs after prompt changes, retriever updates, or model replacements without incurring prohibitive annotation or inference cost.
Despite performing claim-level decomposition, localized evidence matching, and structured verification, RT4CHART remains inexpensive enough to run repeatedly in practice.
This makes it suitable for post-update audits of context faithfulness, in which teams rerun the same evaluation after prompt changes, retriever updates, or model replacements to assess whether outputs have become less well grounded in the retrieved context.

\begin{rqbox}
\textbf{Answer to RQ4:} RT4CHART is robust across judge models and reproducible across repeated runs. Swapping from \resJudgeModel\ to \resAltJudgeModel\ preserves strong performance on both benchmarks, and repeated executions exhibit only marginal variance. Under the default judge model, the cost is approximately \textbf{\$\resCostPerSampleUsd\ per sample}, making routine post-update context-faithfulness auditing practical.
% making routine regression testing RAG auditing practical.}
\end{rqbox}

\subsection{Case Study: Uncovering Overlooked Hallucinations}
\label{sec:case_study_overlooked}

We inspect the \resRagTruthEnhConflictN\ cases where RAGTruth-Enhance disagrees with the original RAGTruth annotations.
In \resRagTruthEnhConflictNoSpanN\ of these cases, the original annotations contain \emph{no} hallucination span, whereas the refined annotations identify at least one.
Across all changed cases, we categorize the \resRagTruthEnhAddedSpanTaxonomyN\ newly added hallucination spans into four recurring patterns, which together account for \resRagTruthEnhAddedSpanMainFourPct\ of the newly added spans:
\begin{itemize}
  \item Unsupported Generalization (\resRagTruthEnhAddedSpanUnsupportedGenN, \resRagTruthEnhAddedSpanUnsupportedGenPct);
  \item Numeric/Logic Inconsistency (\resRagTruthEnhAddedSpanNumericLogicN, \resRagTruthEnhAddedSpanNumericLogicPct);
  \item Inference Stated as Fact (\resRagTruthEnhAddedSpanInferenceFactN, \resRagTruthEnhAddedSpanInferenceFactPct);
  \item Prior Knowledge Interference (\resRagTruthEnhAddedSpanPriorKnowledgeN, \resRagTruthEnhAddedSpanPriorKnowledgePct);
  \item Others (\resRagTruthEnhAddedSpanOthersN, \resRagTruthEnhAddedSpanOthersPct).
\end{itemize}
Below, we present one representative example for each pattern.

\begin{itemize}
    \item \textbf{Numeric/Logic Inconsistency (Sample \#879).}
    The context states that an illness affected ``\textbf{100 people}'' (95 passengers and 5 crew).
    The model rewrites this as ``\textbf{100 passengers} and 5 crew.''
    Although the numbers appear similar, the semantic categories conflict: the source gives 100 total people, not 100 passengers.
    This is a minor arithmetic/logical inconsistency that the original annotations failed to mark.

    \item \textbf{Prior Knowledge Interference (Sample \#587).}
    The context explicitly names the plaintiff as ``\textbf{Joe} Doe,''
    but the model outputs ``\textbf{John} Doe.''
    This error is likely driven by parametric prior knowledge of the common placeholder name rather than by the retrieved evidence.

    \item \textbf{Unsupported Generalization (Sample \#25).}
    The refined annotation adds a baseless span claiming that a country is a ``\emph{common route}'' to another destination for joining an organization,
    while the context states only that it is the \emph{easiest} place to enter the destination.
    The model thus generalizes beyond the retrieved evidence, and Lettuce predicts no hallucination span for this case.\footnote{RAGTruth: no spans (sample id 25). RAGTruth-Enhance adds ``had purchased a visa to Turkey, a common route to Syria for joining the terrorist group.'' (265--353). Lettuce predicts zero spans.}

    \item \textbf{Inference Stated as Fact (Sample \#57).}
    The source presents conditional legal reasoning about whether people associated with the University of Virginia and Phi Kappa Psi \emph{could} sue \emph{Rolling Stone} for defamation.
    The model, however, rewrites this tentative analysis as a settled fact.
    For example, it claims that Phi Kappa Psi's ability to sue is ``\textbf{limited due to the fact that it is not a public figure},'' reversing the source's logic, and further states that the fraternity ``\textbf{cannot demonstrate actual financial harm},'' although the source only says that damages would need to be established.
    The error is therefore not a topic mismatch, but the conversion of tentative reasoning into definite factual claims.
\end{itemize}

These cases reveal a recurring source of benchmark noise: micro-hallucinations that remain superficially plausible while violating exact entities, quantities, or evidential limits.
They further motivate span-level, evidence-anchored evaluation for context-faithfulness auditing.

%% file: 07Limitations.tex
\section{Limitations and Threats to Validity}
\label{sec:discussion}

\subsection{Limitations}
\label{sec:limitations}

\paragraph{Scope: context-only faithfulness.}
RT4CHART evaluates faithfulness strictly with respect to the provided context $C$ (Section~\ref{sec:prelim}).
If $C$ is incomplete, noisy, or itself incorrect, RT4CHART may flag factually correct statements as baseless, or fail to detect errors that are consistently supported by incorrect context passages.
This is an intentional design choice for auditing RAG pipelines, but it limits applicability to settings where the desired notion of truth is grounded in $C$.

\paragraph{Current failure modes.}
RT4CHART can still make errors in two recurring situations.
First, it may miss shifts in certainty, treating hedged evidence and stronger factual restatements as equivalent when their topical content overlaps.
Second, it may over-flag faithful answers when the retrieved context is redundant, internally conflicting, or contains minor surface-form inconsistencies.
Representative examples are included in our artifact~\cite{anonymous_2026_19249142}, highlighting current limitations in handling epistemic nuance and contextual inconsistency.

% RT4CHART can still make errors in two recurring situations.
% First, it may miss shifts in certainty, treating hedged evidence and stronger factual restatements as equivalent when their topic content overlaps.
% Second, it may over-flag faithful answers when the retrieved context is itself redundant, conflicting, or contains minor surface-form inconsistencies.
% We put the failure cases of RT4CHART in our artifact~\cite{rt4hallucination_repo}.
% These cases suggest that RT4CHART remains limited in handling epistemic nuance and context inconsistency.

\paragraph{Residual performance gap.}
Although RT4CHART consistently outperforms the compared baselines, its F1 scores remain far from perfect, especially for fine-grained localization and evidence grounding.
This indicates that context-faithfulness auditing remains an open problem.
Accordingly, RT4CHART should be viewed as an auditing aid for prioritizing suspicious outputs and surfacing evidence-grounded diagnostics, not as a substitute for exhaustive human review in high-stakes settings.

\subsection{Threats to Validity}
\label{sec:threats}

\paragraph{Model and implementation dependence.}
RT4CHART depends on the LLMs used for claim decomposition and verification, and its outputs may vary with decoding behavior and implementation choices.
We mitigate this risk by using deterministic decoding where possible, evaluating cross-model robustness (Table~\ref{tab:rt4chart_glm_stability}), and releasing prompts and code.
In addition, some baselines require adaptation to the $(C,Q,A)$ contract, which may not fully preserve their original operating assumptions.

\paragraph{Generality.}
Our evaluation is based on RAGTruth-derived datasets spanning English-language QA, summarization, and data-to-text tasks.
While this provides diversity in evidence structure and generation style, it remains a bounded evaluation setting.
Accordingly, transferability to non-English corpora, highly specialized domains, or substantially different RAG pipeline designs remains to be established.

%% file: 05RelatedWork.tex
\section{Related Work}
\label{sec:related}

We position RT4CHART along three dimensions: hallucination detection, claim decomposition and fact verification, and test-oracle techniques.

\paragraph{Hallucination detection in LLM outputs.}
Existing detectors~\cite{ji2023hallucination} span several paradigms:
\emph{zero-resource} self-consistency methods~\cite{manakul2023selfcheckgpt,yang2025metaqa},
\emph{model-based} classifiers~\cite{ravi2024lynx,tang2024minicheck,kovacs2025lettucedetecthallucinationdetectionframework}, and
\emph{NLI-based} scorers~\cite{laban2022summac,zha2023alignscore,bowman2015large}.
These approaches typically return holistic scores or labels and do not provide claim-level, context-side evidence under a strict retrieved-context assumption.
RT4CHART differs in that it requires no task-specific training data or response sampling and produces claim-level verdicts along with localized answer spans and context-side evidence.

\paragraph{Claim decomposition and fact verification.}
The decompose-then-verify paradigm---breaking generated text into atomic claims before verification---was established by FActScore~\cite{min2023factscore} and extended by
SAFE~\cite{wei2024longform}, VeriScore~\cite{song2024veriscore}, and
ALCE~\cite{gao2023alce} with search-augmented or citation-based verification.
These approaches share the decomposition structure with RT4CHART but mostly rely on external retrieval or open-world knowledge rather than a fixed retrieved context.
RT4CHART verifies claims strictly against $C$ and does so hierarchically, combining local chunk-level screening with global full-context re-verification (Section~\ref{sec:local-verification}).

\paragraph{Test-oracle techniques in software engineering.}
RT4CHART addresses the \emph{test oracle problem}~\cite{barr2015oracle}: determining correctness when no direct gold output is available.
Classical strategies such as differential~\cite{mckeeman1998differential}, metamorphic~\cite{chen2018metamorphic,segura2016survey}, and intramorphic testing~\cite{rigger2022intramorphic} are either impractical for RAG pipelines (requiring multiple implementations or white-box access) or do not map output claims back to input-side evidence.
\emph{Retromorphic testing}~\cite{yu2023retromorphic} composes a forward program with a backward program and checks consistency in the input modality (Section~\ref{sec:rt-foundation}).
To our knowledge, RT4CHART is the first instantiation of retromorphic testing for RAG faithfulness detection: the RAG pipeline serves as the forward program, and the backward program decomposes the answer into claims, verifies them against the retrieved context, and checks whether the resulting evidence-grounded trace is consistent with~$C$.

%% file: 06Conclusion.tex
\section{Conclusion}
\label{sec:conclusion}

We presented \textbf{RT4CHART}, a retromorphic testing framework for context-faithfulness detection in RAG.
By decomposing answers into independently verifiable claims and enforcing a strict context-only evidence requirement,
RT4CHART moves hallucination detection beyond holistic answer-level scoring toward evidence-grounded, fine-grained diagnosis.
It produces claim-level verdicts, localized answer spans, and context-side evidence, enabling more transparent auditing of grounded generation.
Evaluation on RAGTruth++ and the newly constructed \textbf{RAGTruth-Enhance} shows that RT4CHART consistently outperforms the baselines on answer-level detection and achieves the best span-level answer localization performance.
Our re-annotation results further suggest that standard benchmarks may substantially underestimate hallucination prevalence: RAGTruth-Enhance uncovers \resRagTruthReannotatedHalluMultiplier$\times$ more hallucination cases and \resRagTruthReannotatedSpanMultiplier$\times$ more hallucination spans than the original labels.

% \paragraph{Future work.}
% Several directions follow naturally.
% First, RT4CHART's evidence spans and claim-level labels could support \emph{automated repair}, selectively rewriting only unfaithful claims while preserving faithful ones.~\cite{madaan2023selfrefineiterativerefinementselffeedback}
% Second, improving sensitivity to modal and certainty shifts could reduce false negatives where topical overlap masks changes in epistemic strength.
% Third, handling internally redundant or conflicting retrieved context more robustly may reduce false positives caused by context inconsistency.
% Finally, extending the framework to multi-hop and cross-document settings would test how retromorphic verification scales beyond a single fixed context.~\cite{yang-etal-2018-hotpotqa}

% % Optional unnumbered sections (use only if required by your venue).
% % \section*{Software and Data}
% % The code, prompts, and data used in this paper are available at \url{https://placeholder-url.com}.
% %
% % \section*{Impact Statement}
% % Discuss broader impacts and potential risks/benefits.

%% file: 08Data.tex
\section*{Data Availability}

% The code, prompts, and data used in this paper are available in an anonymous repository for review~\cite{anonymous_2026_19249142}.

Code and data will be made available upon acceptance.

%% file: 09ACK.tex
\section*{Acknowledgements}

This work has emanated from research jointly funded by Taighde Éireann--Research Ireland under Grant Number 13/RC/2094\_2, and by Genesys Cloud Services, Inc.